\documentclass{article}

\usepackage[final, nonatbib]{neurips_2020}

\usepackage[utf8]{inputenc} 
\usepackage[T1]{fontenc}    
\usepackage{hyperref}       
\usepackage{url}            
\usepackage{booktabs}       
\usepackage{amsfonts}       
\usepackage{amsmath}
\usepackage{bm}
\usepackage{amssymb}
\usepackage{nicefrac}       
\usepackage{microtype}      
\usepackage{lipsum}
\usepackage{graphicx}
\usepackage{svg}
\usepackage{ bbold }
\usepackage{multirow}
\usepackage{xcolor}
\usepackage{enumitem}

\title{Prototypical Region Proposal Networks\\
      for Few-shot Localization and Classification}
\author{Elliott Skomski\thanks{Equal contributors.}, Aaron Tuor\footnotemark[1], Andrew Avila, Lauren Phillips\\ \textbf{Zachary New, Henry Kvinge, Courtney D. Corley \& Nathan O. Hodas}\\
Pacific Northwest National Laboratory\\
\texttt{\{firstname.lastname\}@pnnl.gov}}

\begin{document}
\maketitle

\begin{abstract}
    Recently proposed few-shot image classification methods have generally focused on use cases where the objects to be classified are the central subject of images. Despite success on benchmark vision datasets aligned with this use case, these methods typically fail on use cases involving densely-annotated, busy images: images common in the wild where objects of relevance are not the central subject, instead appearing potentially occluded, small, or among other incidental objects belonging to other classes of potential interest. 
    To localize relevant objects, we employ a prototype-based few-shot segmentation model which compares the encoded features of unlabeled query images with support class centroids to produce region proposals indicating the presence and location of support set classes in a query image. These region proposals are then used as additional conditioning input to few-shot image classifiers. We develop a framework to unify the two stages (segmentation and classification) into an end-to-end classification model---PRoPnet---and empirically demonstrate that our methods improve accuracy on image datasets with natural scenes containing multiple object classes.
\end{abstract}

\section{Introduction}
   
    Metric learning approaches for image classification have shown impressive accuracy in the few-shot setting, where models must learn to detect unobserved classes from few labeled examples. The most successful demonstrations have been on datasets such as mini-ImageNet \cite{finn2017model}, ImageNet \cite{deng2009imagenet}, and Omniglot \cite{lake2019omniglot}---datasets comprised of images typically containing centered, non-occluded objects commonly appearing in the foreground. However, performance of state-of-the-art few-shot classification methods can severely degrade on naturally-occurring, so-called ``busy image'' scenes wherein the object to classify is small or occluded, or accompanied by several other incidental objects from other classes. In the common use case where query images have neither image-level annotations nor any indication of where objects of interest are in the image, the few-shot classification task becomes considerably more challenging.
    This work presents and assesses methods for addressing these limitations by conditioning model predictions on regions of interest for improved classification performance in use cases involving busy, unlabeled query images.

    Our presented methods assume the use of an annotated support set---images that define each new few-shot class---wherein each support image is accompanied not only with labels indicating their corresponding class, but also with an annotation mask indicating the objects of interest that belong to the class in question. However, we also explore the use case in which query images are {\em entirely unlabeled}---no image-level annotations or annotation masks are available---and attempt to gain insight into how this can impact model performance. We address three basic research questions related to few-shot classification with busy images:
    
    \begin{enumerate}[noitemsep]
        \item Can annotation information be leveraged for enhanced classification of busy images?
        \item How crucial are query image localization annotations for few-shot busy image classification?
        \item Can few-shot models be leveraged to generate localization proposals for busy, unlabeled query images?
    \end{enumerate}
    
    To address the first question, we compare the accuracy of few-shot classifiers with and without localization, finding that including localization features is  beneficial to performance, particularly on densely-annotated datasets. Then, in comparing performance with and without query images with region annotations, we find that for datasets containing busy images which benefit most from localization, few-shot classification methods which incorporate localization yield sub-par performance when annotations are not available for query images.
    
    As a solution to localizing and classifying busy, unlabeled query images under the few-shot paradigm, we present an $n$-shot, one-way segmentation network which produces query region proposal masks conditioned on a set of annotated support images. 
    We develop an end-to-end classifier we call Prototypical Region Proposal Networks (PRoPnet), whose first stage generates region proposals for query images given a set of support images with class labels and object annotations, and whose second stage augments the standard ResNet-50 architecture with a four-channel input (color channels plus annotation mask) for localization-conditioned classification.
    
    We conduct localization conditioning experiments 
    on seven datasets
    which span a range of classification difficulties---difficult datasets like Visual Genome which contain images of very busy scenes, as well as those which fall on the other end of the spectrum of difficulty, such as ImageNet, where most images depict only objects from a single class of interest in the foreground. 
    Our experiments show that query image annotation information can make few-shot classification a tractable problem for difficult scenes characteristic of Visual Genome, increasing 5-shot, 5-way accuracy from 59 to 76\%. We further demonstrate PRoPnet's improved 5-shot, 5-way classification for the FSS-1000 and PASCAL-5$^i$ datasets. Together, we conclude that i) localization conditioning can be crucial for adequate performance on busy natural image scenes and ii) two-stage networks like PRoPnet which incorporate generated query region proposals as input to a Prototypical Network are a promising approach to providing this essential conditioning. 

\paragraph{Related Work}
    Given saturation of performance on the standard few-shot benchmarks, recent works have extended few-shot tasks to more difficult settings, e.g. object classes with confounding texture \cite{azad2020texture}, cluttered scene segmentation \cite{michaelis2018one, fortin2019few}, long tailed distributions~\cite{wertheimer_few-shot_2019}, and segmentation of homogeneous object clusters ~\cite{wu2018annotation}.
    Enhancements to Prototypical Networks  (ProtoNets)~\cite{snell_prototypical_2017} in particular have been proposed for difficult few-shot tasks involving, for instance,  inhomogeneous noisy datasets~\cite{fort2017gaussian}, domain adaptation~\cite{Pan_2019_CVPR}, and relation classification in text~\cite{gao2019hybrid}.
    Several works have incorporated localization conditioning into a few-shot classification architecture~\cite{karlinsky2020starnet, lifchitz2020few, lin2020fewshot, selvaraju2017grad, wertheimer_few-shot_2019}. We include comparison with the Few-Shot Localization (FSL) method presented in \cite{wertheimer_few-shot_2019} in our experimental evaluations.
    Numerous contemporaneous papers have presented few-shot semantic segmentation methods~\cite{lin2020fewshot, feyjie2020semi, liu2020prototype, wang2019panet, yang2020new, nguyen2019feature, siam2019amp, michaelis2018one, liu2020crnet, zhang2019canet, zhang2020sg, bhat2020learning}, which differ from our approach along several axes of variation including comparison metric, support set aggregation, feature map backbone, and mask refinement.

%
%
\section{Methods}
    In this section we formulate our region proposal and classification architecture, and the implementation, training regimen, and selected datasets. 
    Before describing our methods for region proposal and localized classification under the few-shot paradigm, we begin with a review of the few-shot problem and the Prototypical Network approach to few-shot classification.
    For the few-shot classification task, we have a {\bf \emph{support set}} ${\mathcal{S} = \bigcup_{k=1}^{c} \{ {\bf x}_{k,1}^{(s)}, \dots, {\bf x}_{k,n}^{(s)} \}}$ containing $n$ labeled examples for each of $c$ classes---``$n$-shot, $c$-way'' in few-shot parlance. Given some unlabeled {\bf \emph{query}} example ${\bf x}^{(q)}$ whose class is unknown, our task is to determine which of the $c$ support classes the query example represents.
    We assume some parametric {\bf \emph{encoder}} function $f$ used to map support and query examples to $\mathbb{R}^d$. Let ${{\bf s}_{k,i} = f({\bf x}_{k,i}^{(s)})}$ be the encoded $i$-th example in $\mathcal{S}$ from the $k$-th class.
    The {\bf \emph{support}} for the $k$-th class, is the matrix of encoded examples for that class, ${{\bf S}_k = \begin{bmatrix}{\bf s}_{k, 1}& \hdots & {\bf s}_{k, n}\end{bmatrix}^\mathsf{T} \in \mathbb{R}^{n \times d}}$.
    
    For all few-shot classification models in this work, we use a Prototypical Network \cite{snell_prototypical_2017} variant with a ResNet-50~\cite{he2016deep} encoder function for classification. The {\bf \emph{centroid}} for the $k$-th class is the mean vector, \nobreak ${{\bf \bar{s}}_{k} =\frac{1}{n}\sum_{i=1}^n {\bf s}_{k, i}}$.
    Squared Euclidean distances between the encoded query,  ${{\bf q} = f({\bf x}^{(q)})}$, and support centroids, ${{\bf d}_k = ||\,{\bf q} - {\bf \bar{s}}_{k}\,||_2^2}$, model a class membership probability distribution, ${\hat{{\bf y}} = \text{Softmax}(-{\bf d})}$. 
    Models are trained by minimizing the predicted negative log likelihood of the query's ground-truth class: ${-\log(\hat{{\bf y}}_k)}$, when ${\bf q}$ belongs to the $k$-th class. 
    
\begin{figure}
                \centering
                \includegraphics[width=1.0\textwidth]{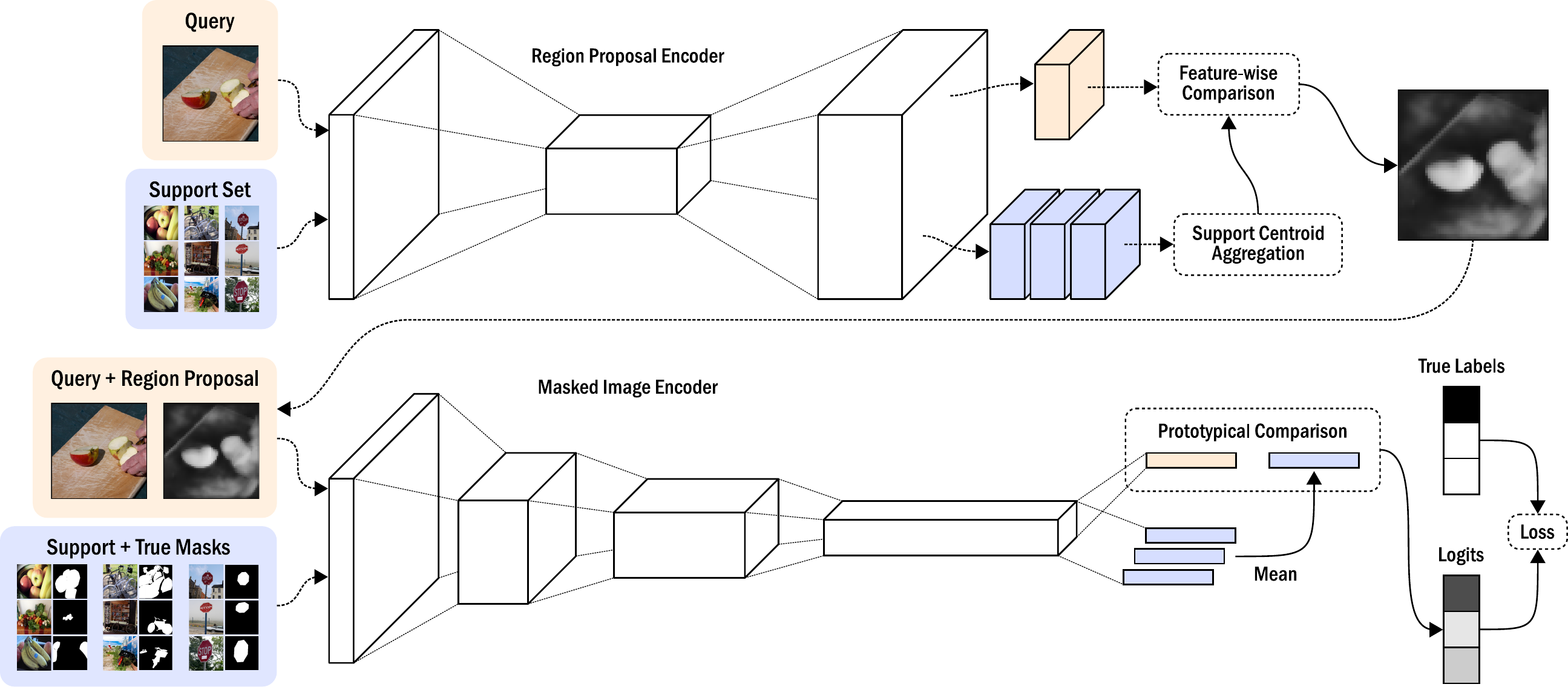}
                \caption{The full end-to-end PRoPnet model.}
                \label{fig:prop}
            \end{figure}

        \paragraph{Region Proposal Network} \label{sec:prop}
            Our region proposal network (RPN) is a logical extension of the prototypical network classification to few-shot segmentation.
            We define a feature map encoder ${\hat{f}: \mathbb{R}^{h \times  w \times 3} \rightarrow \mathbb{R}^{h^{\prime} \times  w^{\prime} \times d}}$ to be a convolutional network that maps an image ${\bf X}$ to a feature map with $d$ channels and possibly downsampled resolution,  where ${h^{\prime} \leq h}$, ${w^{\prime} \leq w}$ and $d > 3$. 
            The RPN is defined as a function which maps a tuple of $n$ support images ${\mathcal{S} = ({\bf X}_1, ..., {\bf X}_n) \in \mathbb{R}^{n \times h \times w \times 3}}$, along with their respective $n$ support annotation masks ${\mathcal{M} =  ({\bf M}_1, ..., {\bf M}_n) \in \{0, 1\}^{n \times h \times w}}$, and a query image ${{\bf Q} \in {R}^{h \times w \times 3}}$ to a query mask ${{\bf P} \in [0, 1]^{h \times w}}$. Specifically, with $i,j$ indexing vertical and horizontal pixel positions, and $k$ indexing the support set exemplars:
            \begin{subequations}
            \begin{align}
                \text{RPN}(\mathcal{S}, \mathcal{M}, {\bf Q})_{i, j} =\:& 
                s_{\text{cos}}\bigr({\bf c}, \hat{f}({\bf Q})_{i, j, :}\bigl)\\
                {\bf c} =\:& \frac{1}{n} \sum_{k=1}^n {\text{MAP}\bigl({\bf M}_k, \hat{f}({\bf X}_k)\bigr)}\\
                \text{MAP}({\bf M}, \hat{f}({\bf X})) =\:& \frac{\sum_{i=1}^h \sum_{j=1}^{w} m_{i, j} \hat{f}({\bf X})_{i, j, :}}{\sum_{i=1}^h \sum_{j=1}^{w} m_{i, j}}
            \end{align}
            \end{subequations}
            
            where ${\bf c}$ is a support class centroid derived by averaging the Masked Average Pooled (MAP) feature representations for the support set class given masks ${\bf M} \in \{0, 1\}^{h \times w}$ and feature maps $\hat{f}({\bf X})$. 
            Our query mask prediction is then derived by a pixel-wise cosine similarity, $s_{\text{cos}}$, comparison  of query pixel features with the class centroid vector. Figure \ref{fig:prop} shows a computational graph of the region proposal network. 
            We choose the feature map encoder, $\hat{f}$, as the portion of UperNet \cite{xiao_unified_2018} preceding their final $1 \times 1$ convolution classification layer. RPN models are trained using the Lov{\'a}sz-Softmax loss~\cite{berman2018lovasz} between the predicted and ground truth query masks.
          
            \paragraph{Prototypical Region Proposal Network} \label{sec:propnet}
            Our Prototypical Region Proposal Network (PRoPnet) for localized few-shot classification is a composition of the RPN and a Prototypical Network modified with early fusion localization conditioning. Query masks are generated by the RPN, then used in a four-channel ResNet-50 feature encoder for ProtoNet few-shot classification (Figure \ref{fig:prop}). Unlike standard ProtoNets, for the classification stage of our network, the query has a different feature representation corresponding to each support class. 
            
    \paragraph{Implementation}
    For the ResNet-50 model used by our localized few-shot classifier, we adapt the reference implementation provided by Torchvision\footnote{\url{https://github.com/pytorch/vision/blob/master/torchvision/models/resnet.py}}.  We add an option to their ResNet-50 to replace the input layer with a four-channel convolutional layer. CSAIL's UperNet implementation\footnote{\url{https://github.com/CSAILVision/unifiedparsing}} was used for our region proposal model---we retain only the Object head of the UperNet for prototypical segmentation. For the FSL baseline \cite{wertheimer_few-shot_2019}, we integrate their few-shot localization code\footnote{\url{https://github.com/daviswer/fewshotlocal}} into our test harness---we omit the covariance pooling and batch folding techniques from our study.
    
    \subsection{Training}\label{sec:train} 
            For the end-to-end system, we employ a two-stage training regimen. In the first stage of training the RPN and early fusion classifiers are trained independently. The RPN is initially trained for few-shot segmentation on Open Images. The early fusion ResNet-50 classifier is independently trained with ground truth masks as a standard image classifier, and then trained on the few-shot classification task, once again with ground truth masks.  
            For the second stage of training, the final layers of the early fusion classifier are fine-tuned to adapt to masks generated from the RPN.
            Each epoch of few-shot training runs a given number of episodes in which a random combination of classes is selected and split into support and query classes, then examples for these classes are sampled to form proper support and query sets. For our experiments, we perform 500 sampling episodes per epoch for 100 epochs during training, and validate our models with 100 sampling episodes.
            We use data augmentation during training by applying random horizontal flipping, rotations, translations, and uniform scaling to both images and corresponding masks. 
        
    \subsection{Datasets}\label{sec:data} Seven publicly available image datasets were selected which meet the criteria of a large number of classes for few shot training and evaluations, and object class label and localization annotations. We selected ILSVRC 2012 (ImageNet)~\cite{russakovsky2015imagenet} and iNaturalist~\cite{van2018inaturalist} as these results can be easily benchmarked against previous few-shot research. Visual Genome (VG)~\cite{krishnavisualgenome} was selected as an exemplary example of a difficult dataset containing many busy real world scenes. We select four semantic segmentation datasets: COCO~\cite{lin2014microsoft}, FSS-1000~\cite{li2020fss}, PASCAL VOC~\cite{everingham2010pascal}, and Open Images~\cite{kuznetsova2020open}. COCO, FSS-1000, and VOC have all been featured as experimental datasets in recent work on few-shot localization. Recently becoming the \emph{de facto} standard benchmark for few-shot semantic segmentation, we use a version of VOC, PASCAL-5$^i$ \cite{shaban2017one} which designates four folds for cross-validation, each containing 15 classes for training and 5 for testing. 
    
    We filter classes in VG, iNaturalist, and Open Images to those which contain at least 200 example images, and exclude object annotations which cover less than 0.2\% of the image. We combine the standard alias classes listed for VG and remove annotations for spurious class definitions such as ``air,'' ``the,'' ``a,'' etc. which do not correspond to a homogeneous class of physical objects. 
    After filtering, we create random train/validation/test splits over classes. For VG, iNaturalist, ImageNet, and Open Images which have a large number of classes, we perform an 80/10/10 split over classes. Since few-shot training and validation involve sampling combinations of classes, for COCO which has only 80 classes, we split the classes 60/20/20 such that train contains 48 classes and validation and test each contain 16 classes.
    For FSS, we use the test split provided by the dataset authors, and derive train and validation classes by forming a validation set with as many classes as the test set, then forming the training set from leftover classes.
    For datasets with multiple classes per image, we remove all test images from the training and validation sets.
    
    Table \ref{tab:dset_stats} shows dataset statistics which indicate relative difficulty of the respective classification tasks. Datasets with more classes and many images per class such as ImageNet have historically been demonstrated to be easier for few-shot tasks. We expect datasets with a high number of classes per image like VG and COCO to benefit most from localization as this statistic is a key indicator for how cluttered typical images are. Mean Area/Sample is the average number of pixels for an annotation, with objects covering less area tending to be more difficult to localize and classify.
    
    \begin{table}
        \resizebox{\textwidth}{!}{
            \renewcommand{\arraystretch}{1.1}
            \centering
            \begin{tabular}{clrrrrr}\toprule
                & {\bf Dataset} & {\bf Samples} & {\bf Classes} & {\bf Imgs/Class} & {\bf Classes/Img} & {\bf Mean Area/Sample} \\ \midrule
                \multirow{4}{*}{\rotatebox[origin=c]{90}{{\bf B-Box}}}
                & {\bf ImageNet} & 593,233 & 1,000 & 593 & 1 & 0.49 \\
                & {\bf iNaturalist} & 282,898 & 530 & 593 & 1 & 0.23 \\
                & {\bf COCO} & 251,855 & 80 & 3,148 & 2.13 & 0.193 \\
                & {\bf Visual Genome} & 1,027,725 & 841 & 1,222 & 9.69 & 0.189 \\ \midrule
                \multirow{4}{*}{\rotatebox[origin=c]{90}{{\bf Segment}}} & {\bf FSS-1000} & 10,000 & 1,000 & 10 & 1 & 0.234 \\
                & {\bf COCO} & 251,855 & 80 & 3,148 & 2.13 & 0.151 \\ 
                & {\bf OpenImages} & 1,261,147 & 345 & 3,656 & 1.41 & 0.172 \\
                & {\bf PASCAL VOC} & 4,318 & 20 & 1,222 & 1.48 & 0.17 \\ \bottomrule \\
            \end{tabular}
            }
            \caption{Summary statistics for each dataset following preprocessing and before splitting. Samples are counted as image-annotation pairs.}
            \label{tab:dset_stats}

        \end{table}
%
%
   
\section{Experiments}

    Although it may be a fair assumption that incorporating localization features should improve a few-shot classifier's accuracy, we first want to verify this empirically and discover whether localization always benefits few-shot models under a variety of conditions. We introduce a systematic analysis of localized few-shot image classifiers combining two axes of comparison.

    \paragraph{Importance of Localized Queries}
    The paradigm few-shot classification task assumes that only support examples are labeled, and query examples are totally unlabeled. Consequently, we assume that typical use cases for few-shot image classifiers with localization will have region annotations for support set images only. However, observing how the presence or absence of annotated query images affects model accuracy can provide insight into whether generating region proposals for query images is a worthwhile endeavor. To evaluate this directly, we train and evaluate few-shot classifiers under two major experimental configurations:
    \begin{itemize}[noitemsep]
        \item {\em Oracle}: use ground-truth annotations for both support and query images,
        \item {\em Support}: use ground-truth annotations for support images only.
    \end{itemize}
    Under the latter configuration, query images are given localizations that encompass the entire image ({\em i.e.} masks comprised entirely of ones).

    \paragraph{End-to-End Model}
    Finally, we evaluate whether few-shot classification performance can be improved upon by incorporating query image localizations generated by a few-shot segmentation model. We train and test our proposed PRoPnet model on a host of vision datasets, testing our architecture on datasets which include either bounding box or segmentation localization features. We compare our approach with the Oracle and Support localization, as well as the method proposed in \cite{wertheimer_few-shot_2019}.

    \subsection{Results and Analysis}
        \begin{table}
            \centering
            \resizebox{\textwidth}{!}{
            \renewcommand{\arraystretch}{1.1}
            \begin{tabular}{lccccccccc}\toprule
            & \multicolumn{4}{c}{Bounding Box} && \multicolumn{4}{c}{Segmentation}\\
            \cmidrule{2-5} \cmidrule{7-10} 
                {\bf Model} &
                {\bf VG} & {\bf COCO} & {\bf iNat} & {\bf ImageNet} &&
                {\bf OpenImg} & {\bf COCO} & {\bf FSS} & {\bf P-5$^i$} \\ \midrule
                {\bf No Localization} &
                    59.27 & 72.31 & 93.79 & 91.50 &&
                    88.73 & 72.44 & 96.21 & 60.79 \\
                {\bf EF Support} &
                    60.22 & 69.71 & 93.09 & 90.59 &&
                    87.67 & 68.84 & 97.02 & 60.15 \\
                {\bf EF Oracle} &
                    76.24 & 82.90 & 94.05 & 93.75 &&
                    92.67 & 84.51 & 98.17 & 73.40 \\ \midrule
                {\bf PRoPnet} &
                    \textcolor{blue}{59.66} & 71.44 & 89.42 & 88.78 &&
                    \textcolor{blue}{88.77} & 71.84 & \textcolor{blue}{97.80} & \textcolor{red}{67.79} \\
                {\bf FSL} \cite{wertheimer_few-shot_2019} &
                    \textcolor{blue}{59.87} & \textcolor{blue}{73.74} & \textcolor{blue}{94.61} & \textcolor{blue}{93.08} &&
                    \textcolor{blue}{90.00} & \textcolor{blue}{75.09} & \textcolor{blue}{96.93} & \textcolor{blue}{62.12} \\ \bottomrule \\
            \end{tabular}
            }
            \caption{5-shot, 5-way test set accuracy (average over 11000 5-query test set episodes) for few-shot classifiers without localization (\textbf{No Localization}), ground truth early fusion localization using support only (\textbf{EF Support}), and using both support and query (\textbf{EF Oracle}), and two methods with localization by region proposal---our proposed method (\textbf{PRoPnet}) and Few-shot Localization (\textbf{FSL})~\cite{wertheimer_few-shot_2019}.}
            \label{tab:cls_with_local_comparison}
        \end{table}
            We report 5-shot, 5-way few-shot classification accuracy on the test set. Accuracy is averaged over 11000 episodes, each with 5 query image predictions. 
            The first three rows of Table \ref{tab:cls_with_local_comparison} give few-shot classification results for models incorporating different degrees of localization information: no localization, localized support and query images (Oracle), or localized support images only (Support).
            Comparing models without localization to those with Oracle localization, for all datasets, we observe accuracy gains via early conditioning on full localization information (Oracle). For datasets with one class per image (ImageNet, iNat, FSS) Oracle localization provides marginal gains of less than 2\% accuracy. However, datasets with multiple classes per image show greater improvement with full localization conditioning. The three datasets with highest average number of classes per image (VG, COCO, P-5$^i$) show drastic improvement with over 10\% gains in accuracy. For these three datasets which contain many crowded scenes, without query localization to indicate which subjects in an image are relevant, the model must learn to localize {\em and} classify, a more difficult learning problem. 
            Comparing Oracle to Support localization, model performance drops without localized query images, \emph{especially} for densely-annotated datasets. This indicates the importance of approaches for localization of query images in the busy image few-shot setting. 
            

            The last two rows of of Table \ref{tab:cls_with_local_comparison} give results for our proposed early fusion annotation conditioning, PRoPnet, and experiments using a late fusion localization conditioning method, FSL, proposed by \cite{wertheimer_few-shot_2019}. Blue colored table entries indicate improvement of a localization method from the No Localization baseline, with the single red colored entry (P-5$^i$) indicating the greatest overall performance improvement. The first four columns show results of the respective localization methods using bounding box annotations. The FSL method is able to leverage this information for more accurate classification whereas PRoPnet does not successfully leverage coarse bounding box support localization, having been designed for pixel-level annotations.
            The last four columns of Table \ref{tab:cls_with_local_comparison} give results on the datasets with fine-grained segmentation annotations.
            Segmentation datasets containing fewer classes and with the largest average classes per image (P-5$^i$ and COCO), demonstrate the largest gains from the localization conditioning methods. FSL improves accuracy on COCO by \textasciitilde{}2.5\% and PRoPnet improves accuracy on P-5$^i$ by \textasciitilde{}7\%. 
            
            
\section{Conclusion and Future Work}
    
    In this work we performed a systematic analysis on few-shot classification with localization using several vision datasets with varying statistical properties. 
    A comparison of classification performance with and without ground-truth localized queries reveals that accuracy declines without localized queries for densely-annotated datasets. However, we find that datasets with one class per image can, to a lesser extent, also benefit from ground-truth localization, likely because less confusion about which objects in an image are relevant obviates any explicit need for localization.

    We present PRoPnet, an end-to-end model for joint few-shot region proposal and classification to improve accuracy on the difficult few-shot task of classifying objects in busy natural scenes. 
    For segmentation datasets, PRoPnet shows improvement over classification with no localization and with ground-truth localized support images for datasets where cluttered natural scenes are commonplace. In addition, datasets with fewer classes benefit most from conditioning on region proposals despite being more challenging in the few-shot context, suggesting that these additional features are more beneficial to highly-constrained few-shot classification tasks.
    Overall, our results suggest that full localization information is essential for classifying objects in cluttered natural scenes, and two-stage networks like PRoPnets and FSL are promising approaches to generate query region proposals which can provide decisive context. 
    
    For future work we plan to extend our two-stage approach to effectively utilize coarser grained annotation information in the form of bounding boxes by replacing the first stage few-shot segmentation architecture with a few-shot object detection architecture. 
    
    

\section*{Acknowledgements}
This work was funded by the U.S. Government. 

\bibliography{egbib}
\bibliographystyle{acm}
    
\end{document}